\documentclass[10pt,twocolumn,letterpaper]{article}

\usepackage{cvpr}              

\usepackage{graphicx}
\usepackage{amsmath}
\usepackage{amssymb}
\usepackage{booktabs}
\usepackage{makecell}
\usepackage{algorithm}
\usepackage{algorithmic}
\usepackage{mathtools}
\usepackage{array}
\usepackage{xcolor}
\usepackage{booktabs}
\usepackage{amssymb}
\usepackage[accsupp]{axessibility}

\usepackage[pagebackref,breaklinks,colorlinks]{hyperref}

\usepackage[capitalize]{cleveref}
\crefname{section}{Sec.}{Secs.}
\Crefname{section}{Section}{Sections}
\Crefname{table}{Table}{Tables}
\crefname{table}{Tab.}{Tabs.}

\def\cvprPaperID{3895} 
\def\confName{CVPR}
\def\confYear{2023}

\begin{document}


\title{$\mathrm{\bf S^3C}$: Semi-Supervised VQA Natural Language Explanation via \\ Self-Critical Learning}
\author{Wei Suo$^{1}$\footnotemark[1] , Mengyang Sun$^{2}$\footnotemark[1] , 
Weisong Liu$^{1}$, Yiqi Gao$^{1}$, Peng Wang$^{1}$\footnotemark[2] , Yanning Zhang$^{1}$\footnotemark[2] , Qi Wu$^{3}$
\\
$^1$School of Computer Science and Ningbo Institute, Northwestern Polytechnical University, China.\\
$^2$School of Cybersecurity, Northwestern Polytechnical University, China.\\
$^3$University of Adelaide, Australia. \\
{\tt\small \{suowei1994,sunmenmian,liuweisong,gyqjz\}@mail.nwpu.edu.cn
}}

\maketitle
\renewcommand\thefootnote{\fnsymbol{footnote}}
\footnotetext[1]{These authors contributed equally to this work.} 
\footnotetext[2]{Corresponding authors.} 
\renewcommand{\thefootnote}{1}
\begin{abstract}
VQA Natural Language Explanation (VQA-NLE) task aims to explain the decision-making process of VQA models in natural language. Unlike traditional attention or gradient analysis, free-text rationales can be easier to understand and gain users' trust. Existing methods mostly use post-hoc or self-rationalization models to obtain a plausible explanation. However, these frameworks are bottlenecked by the following challenges: 1) the reasoning process cannot be faithfully responded to and suffer from the problem of logical inconsistency. 2) Human-annotated explanations are expensive and time-consuming to collect. In this paper, we propose a new Semi-Supervised VQA-NLE via Self-Critical Learning ($S^3C$), which evaluates the candidate explanations by answering rewards to improve the logical consistency between answers and rationales. With a semi-supervised learning framework, the $S^3C$ can benefit from a tremendous amount of samples without human-annotated explanations. A large number of automatic measures and human evaluations all show the effectiveness of our method. Meanwhile, the framework achieves a new state-of-the-art performance on the two VQA-NLE datasets. 
\end{abstract}
\section{Introduction}
\begin{figure}[t]
\centering
\includegraphics[width=0.45\textwidth]{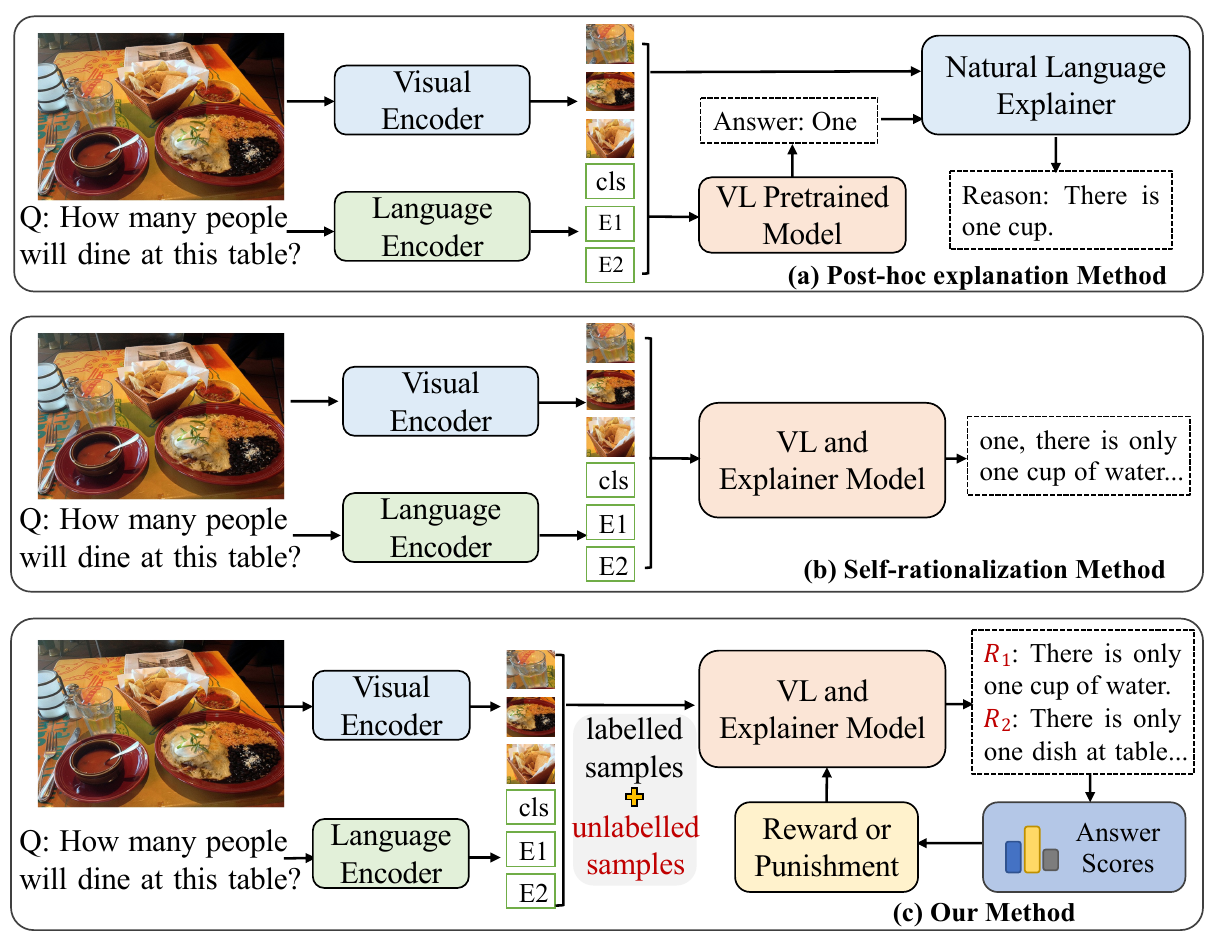}
\caption{Paradigm comparison of different VQA-NLE methods. (a) Post-hoc explanation method adopts two independent models to predict answers and explanations respectively. (b) Self-rationalization method uses a united VL model to simultaneously generate answers and explanations. (c) Our self-critical strategy utilizes answer scores as rewards and obtains more reliable rationales with semi-supervised learning.}
\label{fig:fig1}
\end{figure}
Deep neural networks have enabled significant breakthroughs in a variety of vision-language (VL) tasks such as image captioning~\cite{rennie2017self,cornia2020meshed} and visual question answering (VQA)~\cite{pan2020x,anderson2018bottom}. Unfortunately, most of them are black box systems, which makes it challenging to gain users' trust~\cite{kayser2021vil}. Explaining the decision-making process of deep VL models is a long-standing and essential problem. Some approaches depend on attention mechanisms~\cite{lu2016hierarchical,anderson2018bottom} or gradient-based localization ~\cite{selvaraju2017grad} to acquire visual explanations, which can highlight some contributing image regions for the predicted answers.
However, simple visualization cannot explain how these areas support the answers and they are also hard to comprehend~\cite{kayser2021vil,sammani2022nlx}. Conversely, Natural Language Explanation (NLE) task~\cite{camburu2018snli,narang2020wt5} can explain the decision-making process of a model by generating a natural language sentence. The language-based explanations are more accessible for users to understand, and they can also help researchers optimize the structure of models~\cite{marasovic2020natural}. 

Recently, some models of NLE in the VL community have achieved pretty-well results, especially for VQA-NLE~\cite{kayser2021vil,marasovic2020natural,park2018multimodal,wu2018faithful,sammani2022nlx}. They can guide models to generate natural language sentences and interpret how the models get answers. 
Specifically, the first research line usually treats VQA-NLE as a \emph{predict-then-explain} task~\cite{kayser2021vil,marasovic2020natural,park2018multimodal,wu2018faithful}, namely post-hoc explanations method. As shown in Fig.~\ref{fig:fig1} (a), these methods first depend on pre-trained VL models (such as UNITER~\cite{chen2020uniter} or Oscar~\cite{li2020oscar}) to gain answers. Then the fused multi-modal features and the predicted answers are fed into a separated language model (\emph{e.g.,} LSTM~\cite{hochreiter1997long} or Transformer~\cite{vaswani2017attention}) to generate corresponding explanations. As shown in Fig.~\ref{fig:fig1} (b), the other line~\cite{sammani2022nlx} relies on a united VL model while generating both answers and explanations, which is known as the self-rationalization method. This framework can simultaneously predict an answer and generate a rationale by formulating the answer as a text-generation task along with the explanation.
 
Though significant progress has been made, the two paradigms are still restricted by the following challenges: 
1) For the first paradigm, since the decision-making model and interpretation part are two separate modules, it would inevitably lead to unfaithful responses to the reasoning process of the decision models. 
2) Due to the lack of explicitly logical relationship modeling, previous work ~\cite{jung2022maieutic} has proved that the straightforward self-rationalization frameworks suffer from the problem of logical inconsistency.
3) The above strategies all require an amount of human-annotated explanations, which are expensive and time-consuming to collect~\cite{yordanov2021few}. 

To solve the above challenges, inspired by \cite{sia2022logical,bencivenga2002free}, we argue that a reasonable rationale can assist the model in obtaining a correct answer, and vice versa, the answer can be converted as an evaluation criterion for possible explanations. In this paper, we propose a new \textbf{S}emi-\textbf{S}upervised VQA-NLE method with \textbf{S}elf-\textbf{C}ritical learning, which is called ${S^3C}$ for short. 
As shown in Fig.\ref{fig:fig1} (c), given images and related questions, we first leverage a prompting mechanism to construct answer and explanation templates, which can guide the pre-trained VL model to generate answers and multiple candidate explanations based on sequence sampling~\cite{anderson2018bottom}. 
Then we design a new self-critical method that converts the answer scores as rewards and encourages the model to generate the explanations which contribute to improving the answer scores. In particular, to reduce the dependency on expensive human annotations, we further extend our method to the semi-supervised version, which utilizes the unlabelled samples
\footnote{In this paper, we use ``unlabelled samples'' and ``labelled samples'' to indicate the question-answer (QA) pairs without/with human explanations.}
(\ie, conventional VQA data~\cite{marino2019ok,antol2015vqa}) to significantly enhance the self-interpretability of the model. With the self-critical strategy and the semi-supervised learning, our method effectively models the logical relationships and promotes the logical consistency between answer-explanation pairs. According to automatic measures and human evaluations, the ${S^3C}$ outperforms the state-of-the-art models for the VQA-NLE task on the widely used two datasets and provides a new paradigm for our community. In summary, we make the following contributions:

1) We propose a new self-critical VQA-NLE method that can model the logical relationships between answer-explanation pairs and evaluate the generated rationales by answering rewards. This strategy effectively improves the logical consistency and the reliability of the interpretations.

2) We develop an advanced semi-supervised learning framework for VQA-NLE, which utilizes amounts of samples without human-annotated explanations to boost the self-interpretability of the model further. To the best of our knowledge, we are the first to explore semi-supervised learning on the VQA Natural Language Explanation.

3) The proposed $S^3C$ achieves new state-of-the-art performance on VQA-X~\cite{dong2018predicting} and A-OKVQA~\cite{schwenk2022okvqa} benchmark datasets. Meanwhile, automatic measures and human evaluations all show the effectiveness of our method.

\section{Related work}

\subsection{Explainability in Visual Question Answering}
The visual question answering (VQA) is firstly proposed by~\cite{malinowski2014multi} that requires an intelligent agent to generate an answer by giving an image and a question. Many approaches have been introduced such as joint embedding~\cite{dong2018predicting,YaoPL019Hierachy}, attention mechanisms~\cite{Lu16HieCoAtt,anderson18butd}, memory networks~\cite{XiongMS16,ma18man} and graph neural networks~\cite{GCN,GAT}. Although the VQA task has been well studied, the reasoning process of the models is always agnostic. Some methods apply visualization technologies to achieve visual explanation, such as Grad-CAM~\cite{selvaraju2017grad} and U-CAM~\cite{patro2019u}. However, because image visualization cannot support the answer based on the attended areas~\cite{wu2019self}, in this paper, we focus on improving free-text explanations that are more convenient and easier for users to understand. In this topic, the early work is proposed by~\cite{park2018multimodal}. The paper conducts the VQA-X dataset and utilizes human annotations to inspire the decision-making process of VQA models. 
~\cite{kayser2021vil} designs a new model that combines a pre-trained language model and a VL model to generate free-text explanations. 
~\cite{yang2022chunk} combines stronger pre-trained VL model (\ie Oscar~\cite{li2020oscar}) and generation model (\ie GPT-2~\cite{radford2019language}) to obtain better results.
Recently, ~\cite{sammani2022nlx} proposes a unified model which can simultaneously predict answers and explanations based on a pre-trained caption model. Unlike previous methods, we introduce a new self-critical strategy to model the logical relationships between answers and rationales. It can encourage the model to enhance logical consistency and generate more reasonable explanations.

\subsection{Pre-trained models and Prompt learning}
Pre-trained models have been applied in many fields, such as various NLP tasks~\cite{radford2019language,devlin2018bert} and VL tasks~\cite{tan2019lxmert,lu2019vilbert,zhou2020unified}. Most of these pre-trained models utilize a stack of Transformer structures as the backbone. 
To generalize the pre-trained models to other downstream tasks, previous works mostly fine-tune whole models on each downstream VL task. However, the ability of pre-trained models would be limited due to the mismatch between pre-trained tasks and downstream tasks. Hence, prompt learning~\cite{radford2021learning,zhou2022conditional,jin2022good,cho2021unifying} is proposed, and it can keep the optimization consistency. 
~\cite{radford2021learning} designs the templates to transfer the knowledge to downstream tasks. 
~\cite{cho2021unifying} uses a textual generation framework for uniform optimization. In this paper, we use the pre-trained model based on the image caption task as our backbone. It has been proved that these additional pre-train tasks (\emph{e.g.,} image feature regression or mask language tokens) cannot significantly improve the performance for explanations~\cite{sammani2022nlx}. Meanwhile, we apply different language prompt templates to motivate the model to generate corresponding answers and rationales.
\subsection{Semi-supervised learning}
The development of deep learning depends on a large number of labelled data, while there are many cases in that only a small amount of data can be obtained~\cite{qi2020small}. To solve this challenge, Semi-supervised learning is proposed~\cite{qi2020small}, which aims to train models using a small number of labelled data and amounts of unlabelled data. For example, ~\cite{kingma2014semi} proposes a generatively semi-supervised framework based on variational autoencoders, and it can jointly optimize the model and variational parameters. 
Recently, ~\cite{zhai2019s4l} simultaneously leverages self-supervised and semi-supervised learning to address image classification tasks. 
To our knowledge, our work is the first semi-supervised learning framework for the VQA-NLE task, which effectively alleviates the reliance on expensive human annotations and further boost the self-interpretability of the model.

\section{Method}
\begin{figure*}[t]
  \centering
  \includegraphics[width=1\textwidth]{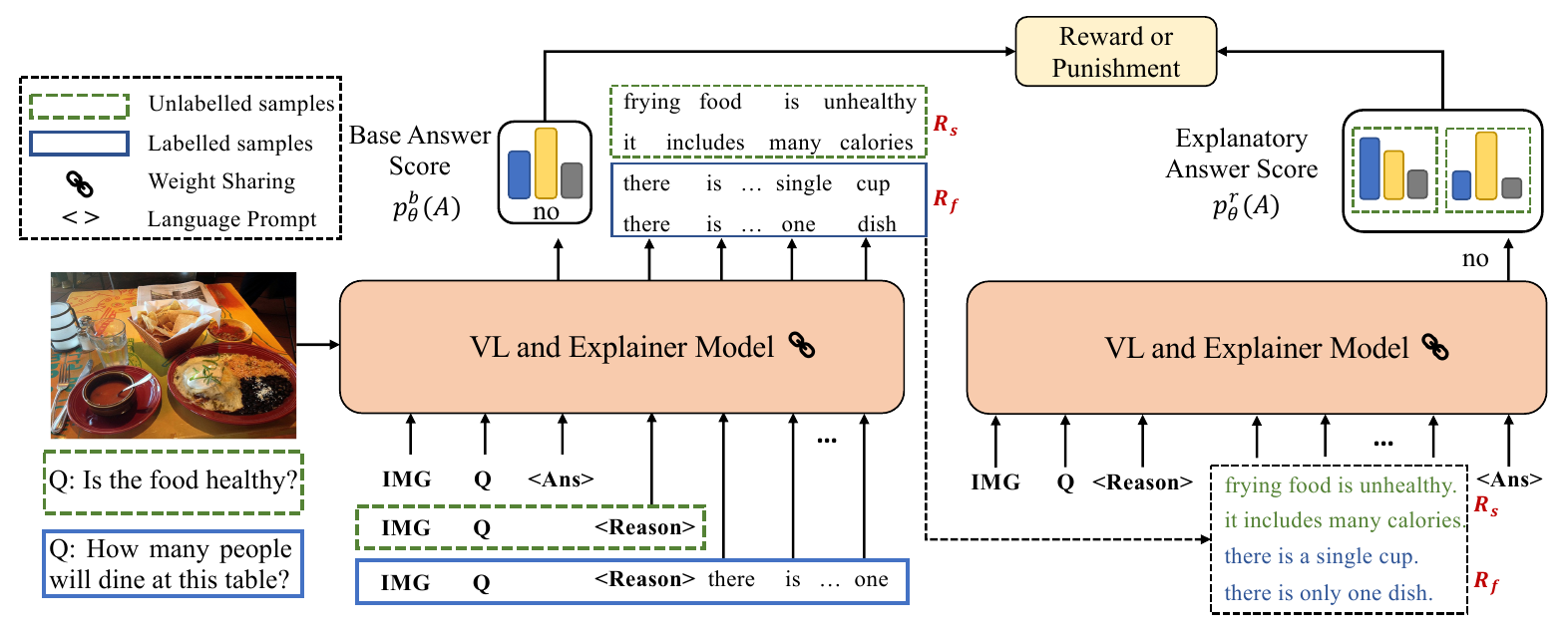}
  \caption{
Overview of our Semi-Supervised VQA-NLE method via Self-Critical learning (${S^3C}$) framework. Given images and corresponding questions for labelled and unlabelled samples, we first use Answer-Explanation Prompt module to obtain the base answer scores and candidate explanations with a pre-trained VL model. Then these reasons are reorganized and fed back into the model to capture the explanatory answer score. Further, our Self-Critical Reinforcement module evaluates the generated explanations and returns the rewards to improve the self-interpretability of the model.}
  \label{fig:fig2}
\end{figure*}

In this section, we introduce our Semi-Supervised VQA-NLE method with the Self-Critical learning ($S^3C$) framework. Our aim is to strengthen the logical consistency between answer-explanation pairs and improve the reliability of the rationales. As shown in Fig.~\ref{fig:fig2}, the $S^3C$ comprises an ``Answer-Explanation Prompt'' module and a ``Self-Critical Reinforcement'' module. Unlike previous approaches, our method first uses a prompting mechanism to generate answers and candidate explanations. Then, we design a new self-critical module that converts the answer scores as rewards to evaluate these reasons. Furthermore, this strategy can conveniently apply the unlabelled QA pairs to enrich training data and enhance the self-interpretability. Next, we describe the components of our model in detail.
\subsection{Pre-trained Vision-Language Backbone}
Given an image $I\in{\mathbb{R}}^{{W}\times{H}\times3}$ and a natural language question $Q=\{q_t\}_{t=1}^{T}$, where $q_t$ represents the $t$-th word, $T$ is the length of the question and ${W}\times{H}\times3$ denotes the size of the image. Our goal is to predict the answer and generate a corresponding free-text rationale. Following previous works~\cite{sammani2022nlx,schwenk2022okvqa}, we adopt the CLIP vision encoder~\cite{radford2021learning} and a pre-trained image caption model (\ie, ClipCap~\cite{mokady2021clipcap}) as our basic backbone. During pre-training, the VL model uses image embeddings from the CLIP as prefixes and fine-tunes a language model (\emph{i.e.,} GPT-2~\cite{radford2019language}) to generate image captions. We refer readers to ~\cite{mokady2021clipcap} for more information about the pre-trained model. In this paper, we consider the image $I$ and question $Q$ as the prefixes of the answer-explanation sequences. Specifically, we first apply ViT-B~\cite{dosovitskiy2020image} and ``classification token'' from the CLIP to obtain the image features. Then, a group of light and simple Multi-Layer Perceptron (MLP) is utilized for transforming the image features to the $V=\{v_s\}_{s=1}^{S}$, $v_s\in{\mathbb{R}}^{c}$, where the dimension size $c=768$ and the image sequence length $S=10$. The above computations can be formulated as follows:
\begin{equation}
{v}_1,{v}_2,\cdots,{v}_S = {\rm MLP}({\rm CLIP}(I)).
\end{equation}
Note that we only update the mapping network (\ie, MLP) during training, while the original visual encoder parameters from the CLIP would remain frozen. For question $Q$, each word $q_t$ would be mapped to the corresponding word embedding $e_t\in{\mathbb{R}}^{c}$ by the pre-trained caption model. Finally, we obtain the image and question sequences $Z$:
\begin{align}
Z = [\overbrace{v_1,v_2,\cdots,v_S}^{\rm{image} \atop \rm{embedding}}, \underbrace{e_1,e_2,\cdots,e_T}_{\rm{question}\atop \rm{embedding}}],
\end{align}
where 
$Z$ is the concatenated multi-modal prefixes.

\subsection{Answer-Explanation Prompt Module}
It has been proved that the prompting mechanism can maintain the same optimization objectives between pre-trained and downstream tasks~\cite{liu2022declaration,jin2022good}. Considering the convenience and explainability, we leverage hand-crafted prompts as templates to enable the model to generate answers or explanations.
As shown in Fig.~\ref{fig:fig2}, the $S^3C$ includes two different kinds of templates, which are introduced respectively as follows. 

\textbf{Base Answer Template.}
The core idea of our $S^3C$ is to use the answer scores as evaluation criteria. Hence, we first establish a basic answer template to acquire base answer scores. Different from previous works~\cite{kayser2021vil,park2018multimodal}, our method treats the visual question answering task as a generation task so that the model can produce answers without a predefined answer space. Specifically, for a given image and question sequence $Z$, we use natural language tokens ``the answer is'' to inspire the model to generate proper answers. By concatenating the language prompt, the base answer template $Z_a=[Z;\langle answer \rangle]$ can be  obtained, where $[;]$ and $\langle answer \rangle$ indicate concatenation operation and the specific language prompts, respectively. During training, the base answer template and ground-true answer label $A=\{a_n\}_{n=1}^{N}$ are fed into the VL model, where $a_n$ is the $n$-th answer token and $N$ denotes the length of the answer label. Next, we compute the answer loss conditioned on the $Z_a$ in an autoregressive fashion:
\begin{align}
\label{eq:3}
L_a = -\frac{1}{N} \sum_{n=1}^{N} logP(a_n | Z_a,a_1,a_2,\cdots,a_{n-1};\theta),
\end{align}
where $\theta$ denotes the parameters of the VL model. Moreover, based on the indexes of ground-true answers, we can acquire average probability $p_{\theta}^{b}(A)$ as our base answer score.

\textbf{Explanation Generation Template.}
To produce a reasonable rationale, we leverage the language prompt ``the reason is'' to motivate the model to generate free-text explanations. Like the base answer template, an explanation generation template $Z_e = [Z;\langle reason \rangle]$ is constructed, where $\langle reason \rangle$ is the natural language tokens. For labelled samples, we follow Eq.~\ref{eq:3} in the autoregressive fashion and the cross-entropy loss to compute explanation loss $L_e$. Because there are no human-annotated explanations available, we would not compute any loss for unlabelled QA samples.

\subsection{Self-Critical Reinforcement Module}
To gain logically consistent rationales, in this module, we expand the searching space by introducing sequence sampling algorithm~\cite{anderson2018bottom} and generate a set of candidate explanations. Besides, the answer scores are treated as rewards to encourage the model to output more detailed interpretations. It's worth noting that the above operations would be implemented on both labelled and unlabelled samples.  

\textbf{Candidate Explanation Generation.}
Benefiting from the language-based prompting strategy and the pre-trained VL model, our explanation generation template can easily guide the model to produce human-readable sentences. Hence, for unlabelled QA samples, we directly apply the explanation generation template $Z_e$ to generate the candidate rationales. To be more specific, we utilize beam search~\cite{anderson2018bottom} to sample the top-$K$ words from the VL model probability distribution at each time step and maintain these sequences with the highest probability. Then, these generated sentences are integrated into candidate explanations $R_s=\{r_k^{s}\}_{k=1}^{K}$ for each QA pair without human-annotated explanations, where $r^s_{k}$ and $K$ indicate $k$-th rationale and the size of beam search, respectively. Moreover, for labelled samples, a similar mechanism is used to build the corresponding explanation set $R_f=\{r_k^{f}\}_{k=1}^{K}$. The above operations are implemented in all samples for the following reasons: 1) Larger search space. As shown in Fig.~\ref{fig:fig2}, ``frying food is unhealthy'' and ``it includes many calories'' can both be rational explanations for the answer ``no''. The expanded search space provides more possibilities for generating reliable rationales. 2) Avoid overfitting. For labelled samples, although we can depend on human explanations to train the model, these labels are still one-sided and subjective. Using the sequence sampling strategy can prevent the model from overfitting these specific annotations~\cite {anderson2018bottom,rennie2017self}. In the end, because the labelled and unlabelled samples would be integrated into a mini-batch during training, we simplify $R_f$ and $R_s$ into $R=\{r_k\}_{k=1}^{K}$ to indicate the candidate explanations for each sample.

\textbf{Self-Critical Reward.}
Inspired by ~\cite{bencivenga2002free,sia2022logical}, an ideal rationale can help the model to infer the answer better. Based on this insight, we argue that the answer scores can be converted as self-critical rewards to evaluate these candidate explanations. Considering that this is a non-differentiable operation, we adopt reinforcement learning method~\cite{rennie2017self} to achieve end-to-end training. In particular, given a sample and corresponding candidate explanations $R$,
we design a new input template:
\begin{align}
\label{eq:4}
Z_r^{k} = [Z;\langle reason \rangle;r_k; \langle answer \rangle],
\end{align}
where $Z_r^{k}$ denotes the template of adding possible rationale $r_k$. Then, this template is fed back into the model and obtains average probability $p_{\theta}^{r}(A)$ about the answer, namely explanatory answer scores, where $A$ is the ground-truth answer for the sample. 
Meanwhile, we use the average probability $p_{\theta}^{b}(A)$ from the output of the base answer template as the base scores. By applying reinforcement learning, the gradient is calculated by:
\begin{align}
   \nabla _\theta L_r(\theta) = -\frac{1}{K} \sum\limits_{k=1}^{K}(p_{\theta}^{r}(A)-p_{\theta}^{b}(A))\nabla _\theta log p_\theta (r_k),
\end{align}
where $p_{\theta}(r_k)$ is the probability of $k$-th explanations. Based on the above computation, this gradient would tend to increase the probability of $k$-th rationales when the answer score $p_{\theta}^{r}(A)$ higher than the scores $p_{\theta}^{b}(A)$ from the base answer template. Finally, for labelled samples, we append the human-annotated rationales to the candidate explanation $R$ and predict the answers by cross-entropy loss, namely $L_{ea}$.
\subsection{Loss}
During the process of training, the overall loss function can be represented as follows:
\begin{align}
   L = L_{a}+L_{e}+{L_{ea}}+\lambda{L_{r}},
\label{eq:6}
\end{align}
where $\lambda$ is used for balancing these two different types of losses (\emph{i.e.,} cross-entropy loss and reinforcement loss). At inference time, our model would first generate the rationales about QA pairs, then we use the explanatory template (\ie, Eq.~\ref{eq:4}) to obtain the corresponding answers.


\section{Experiment}
\subsection{Experimental setting}
{\bf Datasets.} Following previous methods~\cite{sammani2022nlx,schwenk2022okvqa}, we mainly carry out the experiments on the two different VQA-NLE datasets: VQA-X~\cite{park2018multimodal} and A-OKVQA~\cite{schwenk2022okvqa}. Meanwhile, since explain annotations are expensive and time-consuming, we also utilize large-scale VQA v2.0~\cite{goyal2017making} and OK-VQA~\cite{marino2019ok} datasets to build the semi-supervised learning paradigm. Next, we will introduce the four datasets.

{\bf VQA-X.} It is a vision-language dataset that provides explanations for justifying the answers. VQA-X is collected from the Visual Question Answering (VQA) dataset~\cite{antol2015vqa} where the images are obtained from the MSCOCO ~\cite{lin2014microsoft}. It consists of 28K images and 33K QA pairs, split into 29K/1.4K/1.9K for training, validation and testing. Meanwhile, VQA-X constructs complementary pairs which provide a question and two semantically similar images with different answers.

{\bf A-OKVQA.} Compared to the VQA-X, the questions of A-OKVQA generally are required commonsense reasoning about the scene described in the images. It includes 24,903 Question/Answer/Rationale triplets, split into 17.1K/1.1K/6.7K for training, validation and testing. It collects images from the COCO 2017~\cite{chen2015microsoft} dataset, and is further filtered to obtain 23.7K unique images. Compared with previous datasets, the A-OKVQA has richer questions and requires broader areas of knowledge for reasoning.

\textbf{VQA \& OK-VQA.}
We use VQA v2.0 and OK-VQA to provide large-scale unlabelled datasets for semi-supervised learning. The VQA v2.0 dataset is widely used for many previous works~\cite{tan2019lxmert,anderson2018bottom}. It consists of 443k questions and 195k images. To select explainable questions instead of some obvious cases (\eg, How many...?, What color...?), we filter out these questions from VQA v2.0 and obtain the $\sim$90k additional questions based on the rules~\cite{park2018multimodal}. On the other hand, we also use knowledge-based OK-VQA~\cite{marino2019ok} dataset to provide unlabelled knowledge-based QA pairs. It contains a total of 14k questions on 14k images.
\begin{table*}
\begin{minipage}{\textwidth}
\begin{minipage}[t]{0.45\textwidth}
\makeatletter\def\@captype{table}
\caption{Comparison with the state-of-the-art methods on the \textbf{VQA-X}. Note that these results are \textbf{unfiltered} scores. $S^{3}C^{\ast}$  denotes the model without unlabelled samples.}
\label{tab:vqax}
\resizebox{1\textwidth}{!}{
\begin{tabular}{c|ccccccc}
\hline
~ &\multicolumn{6}{c}{\textbf{VQA-X}}\\
\hline
Approach & B4 & M & R & S & C & Acc & Human  \\
\hline
CAPS~\cite{park2018multimodal}& 5.9 & 12.6 & 26.3 & 11.9 &  35.2 & 68.6 &-\\
PJ-X~\cite{park2018multimodal} & 19.5 & 18.2 & 43.4 & 15.1& 71.3 & 76.4& 65.4 \\
FME~\cite{wu2018faithful} & 24.4 & 19.5 & 47.7 & 17.9 & 88.8 & 75.5 &-\\
NLX-GPT~\cite{sammani2022nlx}& 25.6 & 21.5 & 48.7 & 20.2 & 97.2& 83.1&70.2 \\
\hline
${S^3C}^{\ast}$ (\rm{ours}) & 26.5 & 22.0 & 49.0 & 20.9 &  100.5& 83.7&73.9 \\
\hline
\end{tabular}}
\end{minipage}\hspace{10mm}
\begin{minipage}[t]{0.48\textwidth}
\makeatletter\def\@captype{table}
\caption{Comparison with the state-of-the-art methods on the \textbf{A-OKVQA}. Note that these results are \textbf{unfiltered} scores. $S^{3}C^{\ast}$ denotes the model without unlabelled samples.}
\label{tab:aokvqa}
\resizebox{1\textwidth}{!}{
\begin{tabular}{c|cccccccc}
\hline
    ~ &\multicolumn{7}{c}{\textbf{AOKVQA}}\\
\hline
    Approach & B4 & M & R & S & C & Acc val & Acc test & Human \\
\hline
    ViLBERT~\cite{lu2019vilbert} & - & - & - & - & - & 30.6 & 25.9 &-\\
    LXMERT~\cite{tan2019lxmert}& - & - & - & - & - & 30.7 &25.9 &-\\
    KRISP~\cite{marino2021krisp} & - & - & - & - & -& 33.7 &27.1 &-\\
    Clipcap~\cite{schwenk2022okvqa} & - & - & -  & -& -& 30.8 &25.9 &-\\
    e-UG~\cite{kayser2021vil} & 15.1 & 18.1 & 42.4 & 14.9 & 51.5 & 30.5 & 25.6 & 44.1 \\
    NLX-GPT~\cite{sammani2022nlx}& 20.1 & 17.0 & 46.3 & 15.8 & 65.4 & 32.7 & 28.7&46.9\\
    \hline
    ${S^3C}^{\ast}$ (\rm{ours}) &21.8 &17.9&47.3& 17.3 &70.6&33.0&29.6&49.4 \\
    ${S^3C}$ (ours) & \textbf{22.5} & \textbf{18.5} & \textbf{48.4} & \textbf{18.1} & \textbf{74.4} & \textbf{34.2} & \textbf{33.5} & \textbf{54.7} \\
\hline
\end{tabular}}
\end{minipage}
\end{minipage}
\end{table*}
\begin{table*}
    \centering
    \caption{Comparison with the state-of-the-art methods on the VQA-X. Note that these results are \textbf{filtered} scores. $S^{3}C^{\ast}$ denotes the model without unlabelled samples.}
    \resizebox{0.7\textwidth}{!}{
    \begin{tabular}{l|cccccccc|ccc}
     \Xhline{1.2pt}
        ~ & B1 & B2 & B3 & B4 &M& R &S & C & Acc & Human \\
        \hline
        RVT~\cite{marasovic2020natural} & 51.9 & 37.0 &25.6 & 17.4 & 19.2 &  42.1 & 15.8&52.5 & 68.6    & 60.5\\
        PJ-X~\cite{park2018multimodal}  & 57.4 & 42.2 &30.9 & 22.7 & 19.7  & 46.0 & 17.1 &82.7 & 76.4   & 69.3\\
        FME~\cite{wu2018faithful}       & 59.1 & 43.4 &31.7 & 23.1 & 20.4 & 47.1 & 18.4& 87.0 & 75.5    & -\\
        QA-only~\cite{kayser2021vil}    & 51.0 & 36.4 &25.3 & 17.3 & 18.6 & 41.9 & 14.9& 49.9 & -       & -\\
        e-UG~\cite{kayser2021vil}       & 57.3 & 42.7 &31.4 & 23.2 & 22.1  & 45.7 & 20.1&74.1 & 80.5    & 71.4\\
        NLX-GPT~\cite{sammani2022nlx}   & 64.2 & 49.5 &37.6 & 28.5 & 23.1 & 51.5 & 22.1&110.6 & 83.2    & 73.7\\
        \hline
        ${S^3C}^{\ast}$ (\rm{ours})     & 64.4 & 49.9 & 38.0 & 29.1 & 23.4  & 51.9 & 22.7&112.1 & 83.7  & 75.9\\
        $S^3C$ (ours)                   & \textbf{64.7} & \textbf{50.5} &\textbf{38.8} & \textbf{30.7} & \textbf{23.9} & \textbf{52.1} & \textbf{23.0}&\textbf{116.7} & \textbf{85.6}& \textbf{79.2} \\      
    \Xhline{1.2pt}
    \end{tabular}}
    \label{tab:filtered}
\end{table*}

\textbf{Implementation Details.} Following~\cite{mokady2021clipcap}, each image is first pre-processed (such as including image resize, center crop, and normalize) by CLIP~\cite{radford2021learning}. Meanwhile, we fix the ViT-B weights from the CLIP visual encoder to accelerate the training speed. For the mapping network, the image sequence length $S=10$ and the embedding size is 768. The AdamW~\cite{kingma2014adam} is used as our optimizer with the weight decay 1e-5, and the batch size and beam size $K$ are set to 4 and 2. The weight coefficient $\lambda$ is set to 10. We train all models on the four 1080Ti GPUs for 30 epochs with a learning rate of 1e-5. 
\subsection{Evaluation Measures}

{\bf\noindent Automatic Metering.} Following~\cite{sammani2022nlx,yang2022chunk}, we use the automatic metrics BLEU~\cite{papineni2002bleu}, METEOR~\cite{denkowski2014meteor}, ROUGE-L~\cite{lin2004rouge}, SPICE~\cite{anderson2016spice} and CIDEr~\cite{vedantam2015CIDEr} to evaluate generated explanations. For the evaluation of predicted answers, we follow~\cite{sammani2022nlx,yang2022chunk} to compute the VQA accuracy.

{\bf\noindent Human Evaluation.} Automatic VQA-NLE measures do not always reflect the correctness and logicality of the explanations~\cite{kayser2021vil,vaideeswaran2022towards}, thus we also build human evaluations. The process is similar to~\cite{kayser2021vil,marasovic2020natural}. Specifically,
for each explanation, three human evaluators are required to decide whether an explanation can justify the answer and select an option (including ``yes, weak yes, weak no and no''). The selection will be mapped to scores ($1$, $\frac{2}{3}$, $\frac{1}{3}$ and $0$). The final scores are computed by averaging among all test samples. Meanwhile, 
these evaluators are asked to choose the reasons for unqualified explanations. We follow~\cite{kayser2021vil} to define three kinds of aspects to evaluate the explanations: irrelevant explanations, insufficient explanations and meaningless explanations.
First, the irrelevant explanations may not match the image, for example, "have a long neck" is a good explanation for the answer "giraffes" when asked "what animals are these?", but the image may display cows. 
Second, the insufficient explanations only describe the image but cannot corroborate the answers. For example, the sentence ``there are some people'' does not sufficiently justify the question ``do people have a party?''. 
Lastly, some nonsensical sentences could be judged as contradictory explanations, such as “a man is a woman”. 
For each sample, the human evaluators can select multiple shortcomings. More details of the human evaluation can be found in the~\cite{kayser2021vil}.
\begin{table*}
\begin{minipage}{\textwidth}
\begin{minipage}[t]{0.48\textwidth}
\makeatletter\def\@captype{table}
\caption{\textbf{Main shortcomings.} The main shortcomings of unqualified explanations on the VQA-X dataset. For each sample, human evaluators can select multiple shortcomings.}
\label{tab:weak}
\resizebox{1\textwidth}{!}{
\begin{tabular}{lcccc}
\Xhline{1.2pt}
Model  &  \makecell[c]{Irrelevant\\ explanations} & \makecell[c]{Insufficient\\ explanations} & \makecell[c]{Meaningless\\ explanations}    \\
\hline
RVT~\cite{marasovic2020natural} & 25.7\% & 33.5\% & 11.4\% \\
PJ-X~\cite{park2018multimodal} & 21.1\% & 28.4\% & 9.2\% \\
e-UG~\cite{kayser2021vil} & 22.8\% & 25.4\% & 8.7\% \\
NLX-GPT~\cite{sammani2022nlx} & 20.3\% & 22.2\% & 9.1\% \\
\hline
$S^3C$ (ours) & \textbf{17.3\%} & \textbf{18.9\%} & \textbf{8.2\%} \\       
\Xhline{1.2pt} 
\end{tabular}
}
\end{minipage}\hspace{10mm}
\begin{minipage}[t]{0.45\textwidth}
\makeatletter\def\@captype{table}
\caption{\textbf{Cross-dataset testing.} We alternately use the VQA-X and A-OKVQA as source dataset and target dataset to test the generalization of our framework.}
\label{tab:zero-short}
\resizebox{1\textwidth}{!}{
\begin{tabular}{c|cccccc}
     \Xhline{1.2pt}
        ~ &\multicolumn{6}{c}{\textbf{VQA-X}$\rightarrow$\textbf{A-OKVQA}}\\
    \hline
        Approach & B4 & M & R & S & C & Acc   \\
    \hline
        NLX-GPT~\cite{sammani2022nlx} & 10.7 & 12.7 & 34.2 & 10.7  & 35.4 &10.4  \\
        ${S^3C}$ (ours)& \textbf{12.0} & \textbf{13.3} & \textbf{34.3} & \textbf{12.5}& \textbf{45.3} & \textbf{18.8} \\
    \hline
        ~ &\multicolumn{6}{c}{\textbf{A-OKVQA}$\rightarrow$\textbf{VQA-X}}\\
    \hline
        Approach & B4 & M & R & S & C & Acc   \\
    \hline
        NLX-GPT~\cite{sammani2022nlx} &9.1 &13.6 & 32.8 & 9.1 & 33.2 & 42.4 \\
        $S^3C$ (ours)& \textbf{10.9} & \textbf{15.0} & \textbf{34.1} & \textbf{10.4}& \textbf{38.6} & \textbf{43.8} \\
        
    \Xhline{1.2pt}
    \end{tabular}
}
\end{minipage}

\end{minipage}
\end{table*}

\begin{table*}
    \centering
    \caption{\textbf{Ablation study.} We ablate key components to demonstrate the effectiveness of our method. SCR and Semi are Self-Critical Reinforcement module and Semi-supervised learning paradigm respectively.}
    \resizebox{0.85\textwidth}{!}{
    \begin{tabular}{l|cccccc|cccccc}
     \Xhline{1.2pt}
        ~ & question & image & answer & explanation & SCR & Semi & B4 & M & R & S & C &Acc\\
        \hline
        1 & \checkmark & \checkmark & \checkmark & -- & -- & --                      &--&--&--&--&--&80.1\\
        2 & \checkmark & \checkmark & -- & \checkmark & --  & --                     &24.4&20.7&47.3& 19.5& 90.4& -- \\
        3 & \checkmark & \checkmark & \checkmark & \checkmark & --  & --             &27.5&22.9&50.4& 21.9 &109.1& 82.2 \\
        4 & \checkmark & \checkmark & \checkmark & \checkmark & \checkmark  & --     &29.1&23.4&51.9&22.7&112.1 & 83.7\\
        5 & \checkmark & \checkmark & \checkmark & \checkmark & \checkmark & \checkmark    &\textbf{30.7}&\textbf{23.9}&\textbf{52.1}&\textbf{23.0}&\textbf{116.7}&\textbf{85.6}\\
    \Xhline{1.2pt}
    \end{tabular}
    }
    \label{tab:ablation}
\end{table*}

\subsection{Quantitative evaluation.}
\textbf{Automatic Evaluation.}
We compare our method with the state-of-the-art models on the VQA-X and A-OKVQA datasets in Table~\ref{tab:vqax}-\ref{tab:filtered}. The B4, M, R, S, C, Acc and Human are short for BLEU-4, METEOR, ROUGE-L, SPICE, CIDEr, Answer precision and Human evaluation. We use ``unfiltered'' to indicate that the explanations are evaluated regardless of whether the answer is true or false. While ``filtered'' is to only consider the explanations which have correct answers. From Table~\ref{tab:vqax}, the ``${S^3C}$*'' denotes that we only apply the self-critical framework without using unlabelled samples, while the row ``${S^3C}$'' indicates our method in the semi-supervised setting. We observe that the proposed framework outperforms both the post-hoc explanation methods~\cite{park2018multimodal, wu2018faithful} and the self-rationalization method~\cite{sammani2022nlx}. 
Meanwhile, it's worth noting that although the NLX-GPT uses a more powerful pre-trained VL model with larger pre-trained datasets (\ie, COCO captions~\cite{lin2014microsoft}, Flicker30k~\cite{plummer2015flickr30k} and Visual Genome~\cite{krishna2017visual}), our method still obtains 3.3 absolute gains on the CIDEr indicator with fewer pre-trained data (only using COCO captions). Furthermore, when we utilize our proposed semi-supervised paradigm, the results are further improved by 7.2 points. These results show that our model can generate more reliable explanations and it can benefit from the amount of data without human-explanation labels. We also report the accuracy of answers in the column of ``Acc''. It can be observed that our self-critical method with semi-supervised learning can simultaneously boost the precision of answers and corresponding explanations.

In Table~\ref{tab:aokvqa}, we evaluate our method on the A-OKVQA dataset. The results show that the $S^3C$ can outperform all the previous works~\cite{jiang2018pythia,lu2019vilbert,tan2019lxmert,marino2021krisp,schwenk2022okvqa,sammani2022nlx}. Especially, compared to the SOTA model~\cite{sammani2022nlx}, there are 9.0 points and 4.8\% improvements in the CIDEr and answer accuracy based on our model. It demonstrates that our model can benefit from semi-supervised learning and generate explanations about commonsense reasoning. 

In addition, to prove the algorithm's validity, we follow~\cite{sammani2022nlx,kayser2021vil} to report the filtered scores for the VQA-X dataset in Table~\ref{tab:filtered}. Through filtering the correct answers, our method can outperform whatever post-hoc~\cite{marasovic2020natural,park2018multimodal,wu2018faithful,kayser2021vil} and self-rationalization~\cite{sammani2022nlx} methods. Meanwhile, the proposed $S^3C$ achieves a new state-of-the-art with the CIDEr score improved by 6.1 points to 116.7 and boost the answer accuracy by 2.4\% to 85.6.
\textbf{Human Evaluation.} To evaluate the faithfulness and correctness of these generated explanations, we conduct the human evaluation that is shown in the column of ``Human'' in Table~\ref{tab:vqax}-\ref{tab:filtered}. The experimental results further prove our method has better  self-interpretability for the VQA-NLE task. Moreover, we also ask the human evaluators to select the shortcomings for each unqualified explanation on the VQA-X dataset. As shown in Table~\ref{tab:weak}, we build three shortcoming options (\emph{i.e.,} irrelevant explanations, insufficient explanations and meaningless explanations) with a multi-choice process. These results indicate that the ${S^3C}$ can obtain relatively better rationales and empirically confirm the effectiveness of our method.
\begin{figure*}[t]
  \centering
  \includegraphics[width=1\textwidth]{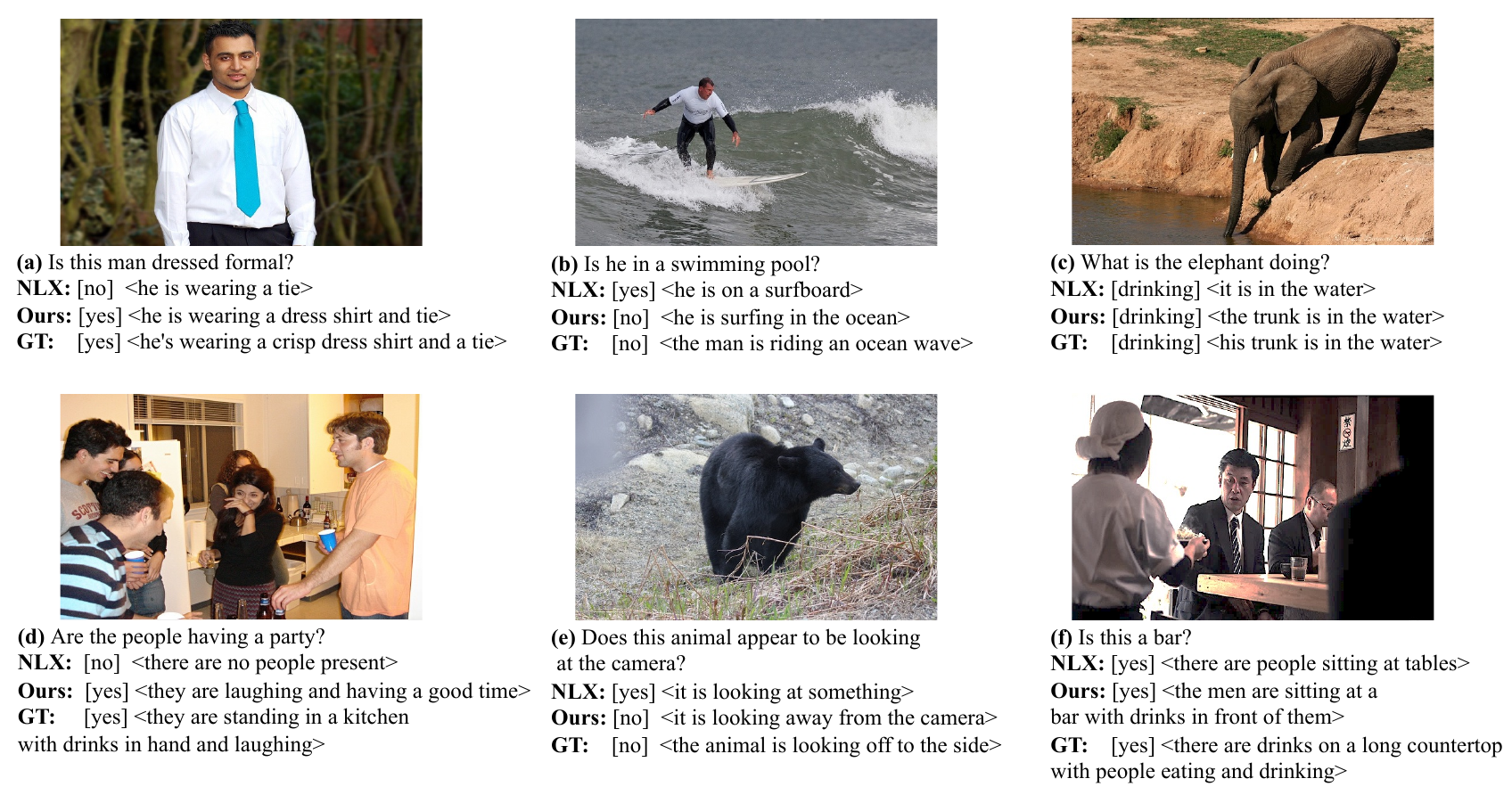}
  \caption{
    Some generation results on the VQA-X datasets. The [ ] and $< >$ indicate answers and explanations respectively.
    We show the results of the state-of-the-art NLX-GPT~\cite{sammani2022nlx}, our method and ground-truth (they are short for NLX, Ours and GT).
  }
  \label{fig:fig3}
\end{figure*}
\subsection{Cross-dataset Performance.}
We propose cross-dataset experiments to measure the generalization and reliability of our framework. For all we know, there is very little research on cross-dataset explanation generation. As shown in the first two rows of Table~\ref{tab:zero-short}, we use the VQA-X and the A-OKVQA as the source dataset and target dataset to test the generalization ability of our framework. Specifically, we use the weights from the state-of-the-art method~\cite{sammani2022nlx}, which are trained on the VQA-X dataset and tested on the A-OKVQA dataset to obtain the results of the NLX-GPT. From the comparisons of six indicators, we find that our self-critical model has better generalization from the general VQA to the knowledge-based VQA. This suggests that ${S^3C}$ can learn commonsense from visual question answering and generate better explanations and answers. Meanwhile, when we exchange the source dataset and target dataset as shown in the last two rows of Table~\ref{tab:zero-short}, our framework can still surpass the existing SOTA model on all evaluation measures. These quantitative results effectively
demonstrate the generalization and stability of our model.

\subsection{Ablation Studies.}
As shown in Table~\ref{tab:ablation}, we conduct several ablation studies on the VQA-X to demonstrate the effectiveness of our method. In particular, we first build the experiments in the 1-3 rows of Table~\ref{tab:ablation}, which use our Answer-Explanation Prompt module with cross-entropy loss as the baseline models. It can be found that when we simultaneously generate both answers and interpretations, the answer accuracy and CIDEr can achieve significant improvements. Meanwhile, these results also prove that the answer-explanation pairs have inherent consistency and they can promote each other. Further, we establish the Self-Critical Reinforcement module as shown in the fourth row, and the result of accuracy and CIDEr are boosted by 1.5\% and 3.0 points. It suggests that our self-critical method with sequence sampling strategy can 
encourage the model to generate more reliable explanations and correct answers. Finally, when we extend our model to the semi-supervised setting, the performance improves to 85.6 on the accuracy and 116.7 on the CIDEr over the baseline model (\emph{i.e.,} the row 5 of Table~\ref{tab:ablation}) by 3.4\% and 7.6 points. The results demonstrate that our framework is remarkably effective in improving the logical consistency and the self-interpretability of the model. 
\subsection{Qualitative evaluation.}
As shown in Fig.~\ref{fig:fig3}, we show some qualitative results from NLX-GPT~\cite{sammani2022nlx} and our $S^3C$ method on the VQA-X dataset. 
Through overall comparison, our model achieves better logical consistency between answers and explanations. For example, in Fig.~\ref{fig:fig3} (a), although the SOTA method~\cite{sammani2022nlx} correctly identifies the significant symbol of formal dress (\ie, tie), the predicted answer is ``no'' that is contradictory to the explanation. On the contrary, our method not only obtains a more complete and faithful rationale but also generates a logically consistent answer. Additionally, the $S^3C$ can also produce more persuasive and trusty sentences. For instance, in Fig.~\ref{fig:fig3} (b)-(f), the generated sentences contain scene description ``having a good time'' and fine-grained rationale  ``the trunk is in the water''.
\section{Conclusion}
In this paper, we propose a new Semi-Supervised VQA-NLE method via Self-Critical Learning (${S^3C}$). Different from previous works, our method first utilizes the prompting mechanism to motivate the model to generate answers and candidate explanations. Meanwhile, we design a novel Self-Critical Reinforcement module, which converts the answer scores as rewards to evaluate these possible rationales. Furthermore, our framework can benefit from an abundance of question-answer pairs without human-annotated explanations and further boost the self-interpretability of the model. With the automatic measures and human evaluations, our ${S^3C}$ achieves a new state-of-the-art on multiple benchmarks and provides a new paradigm for our community.  
\section*{Acknowledgements}
This work was supported by National Key R\&D Program of China (No.2020AAA0106900), the National Natural Science Foundation of China (No.U19B2037), Shaanxi Provincial Key R\&D Program (No.2021KWZ-03), and Natural Science Basic Research Program of Shaanxi (No.2021JCW-03).
{\small
\bibliographystyle{ieee_fullname}
\bibliography{main}
}
\end{document}